\documentclass[11pt,a4paper]{article}
\usepackage[hyperref]{acl2020}
\usepackage{times}
\usepackage{latexsym}
\usepackage{graphicx}
\usepackage{amssymb}
\usepackage{pifont}

\aclfinalcopy 

\usepackage{booktabs}
\usepackage{hyperref}
\usepackage{array}

\usepackage{booktabs}
\usepackage{multirow}
\usepackage{gb4e}  
\usepackage{enumitem}
\noautomath

\usepackage{times}
\usepackage{amsmath}
\usepackage{tipa}
\newcommand{\ipa}[1]{\textipa{#1}}  
\usepackage{CJKutf8}

\usepackage{microtype}

\title{Neural Proto-Language Reconstruction}

\author{Chenxuan Cui \qquad Ying Chen \qquad Qinxin Wang \qquad David R. Mortensen \\
  Language Technologies Institute, Carnegie Mellon University \\
  \texttt{\{cxcui, yingc4, qinxinw, dmortens\}@cs.cmu.edu}}

\date{}

\begin{document}
\maketitle
\begin{abstract}
Proto-form reconstruction has been a painstaking process for linguists. Recently, computational models such as RNN and Transformers have been proposed to automate this process. We take three different approaches to improve upon previous methods, including data augmentation to recover missing reflexes, adding a VAE structure to the Transformer model for proto-to-language prediction, and using a neural machine translation model for the reconstruction task. We find that with the additional VAE structure, the Transformer model has a better performance on the WikiHan dataset, and the data augmentation step stabilizes the training.
\end{abstract}

\section{Introduction}
Languages change over time. An ancestral language, known as the proto-language, can develop into a group of mutually intelligible dialects through time, and eventually diverge into the mutually unintelligible daughter languages of today. Examples of this process include the Romance family, in which Vulgar Latin developed into French, Spanish, Italian, Romanian, and many more; and the Sinitic language family, in which Middle Chinese developed into Mandarin, Shanghainese (Wu), Cantonese (Yue), Hakka, etc.
Linguists understand these language families because a rigorous method is applied to reconstruct the proto-form of languages, which is known as the \textit{comparative method}. The comparative method roughly consists of the following major steps \citep{trask}:

\begin{enumerate}[itemsep=0pt]
\item Gather a large number of cognate sets from (potentially) related languages.
\item Identify sound correspondences among the daughter languages after examining all the data.
\item Reconstruct the proto-form of each cognate set, while ensuring phonetic, phonological and phonotactical validity.
\end{enumerate}

As one can imagine, this is an extremely labor intensive process and does not scale well as the number of daughter languages or the number of cognate sets increase. As a result, a computational model that can partially automate step 2 and 3 of the comparative process would be of great value to the historical linguistics and NLP community. Such a system can work alongside a linguist and improve the efficiency of the reconstruction task.

For this project, we focus on reconstructing the Sinitic family, where the proto-language is Middle Chinese. Currently, prominent reconstructions for Middle Chinese are based on \textit{Qieyun}, a historical rhyme book of the era that records the pronunciation of the elite dialect of China. 
The Sinitic family currently lacks a good comparison-based reconstruction, which was criticized by \citet{norman-coblin}, so this project contributes to the sinology community in this respect. In addition, the Sinitic family also has the advantage of a simpler syllable structure, so reconstruction could be more easily modeled. Finally, a relatively large dataset named WikiHan \citep{chang-etal-2022-wikihan} is recently published, making the task more amenable to neural methods. We will use WikiHan for the majority of the experiments.

In this project, we wish to improve the performance of models on WikiHan. In the rest of this report, we describe three major directions of exploration:
\begin{enumerate}[itemsep=0pt]
\item Use data augmentation techniques to fill in missing entries in WikiHan to improve the performance. (\autoref{sec:data-augmentation})
\item Adding a forward reconstruction model to ensure a more meaningful latent space, resulting in a VAE structure. (\autoref{sec:vae-transformer})
\item Modify a NMT model to suit the language reconstruction task. (\autoref{sec:variational-nmt})
\end{enumerate}


\section{Background} \label{sec:background}

\subsection{Related Work}
The task of proto-language reconstruction has been attempted by the computational linguistics community.
\citet{list-etal-2022-new} has proposed a series of methods based on alignment and classification. In these methods, an alignment algorithm is first used to segment each word in the cognate set into corresponding phonemes. These sound correspondences are then turned into contextual features before an classifier such as SVM is applied to predict the proto-phonemes in each sound correspondence. These methods are fairly naive and do not perform too well. Even on Middle Chinese, where the segmentation portion of the model can be done with a simple rule-based model, the performance is still rather poor, as shown in \autoref{tab:baselines}.

\citet{ciobanu-etal-2020-automatic} is among the first works to formulate proto-language reconstruction as a sequence-to-sequence task. It uses a conditional random field (CRF) along with N-gram features as the model, and achieves fair results on a Romance data. 

One of the first attempts at neural proto-language reconstruction, and one of the state-of-the-art models currently is \citet{meloni-etal-2021-ab}, which uses a character-based LSTM encoder and decoder with attention. As a result, the features are automatically extracted by the RNN, which was not the case for \citet{list-etal-2022-new} or \citet{ciobanu-etal-2020-automatic}. As input to the RNN, the daughter forms are concatenated into a long sequence, with separator tokens in-between to indicate the language ID of the following word. This model achieved great performance in a romance dataset.

An improvement to \citet{meloni-etal-2021-ab} is to replace the LSTM encoder and decoder with a transformer encoder and decoder \citep{11711-report}. The separator tokens to indicate language ID are in turn replaced by language embeddings added to token embeddings and positional embeddings, as in standard transformer architectures. By making this change, the model is more capable of capturing long term dependencies within the input, such as when the daughter forms are long or when the number of daughter languages is large. For example, \citet{11711-report} show that the Transformer model out-performs \citet{meloni-etal-2021-ab} in the romance dataset and in a Sinitic dataset consisting of 39 daughter topolects \citep{Hou2004}, both of which have long concatenated sequences. However, as we will show later in this report, it does not out-perform \citet{meloni-etal-2021-ab} on WikiHan.

For a more comprehensive and detailed overview of the research landscape in proto-language reconstruction, refer to Section 2 of \citet{11711-report}.


\subsection{Evaluation Criteria}
We evaluate the performance of proto-form reconstruction primarily with \textit{accuracy} and \textit{edit distance} (ED). Accuracy is the rate of reconstructed proto-forms that are exactly the same with the reference ones. Edit distance (also known as Levenshtein distance) is the average number of character edits needed to convert the predicted proto-forms to the reference ones. 

There are three more metrics used in the research community that we calculate to supplement the evaluation. \textit{Phoneme error rate }(PER) is the normalized edit distance where each unit is a phoneme, since a phoneme can consist of more than one character. \textit{Feature error rate }(FER) is the normalized edit distance where each unit is a phonological feature. This captures the fact that certain phoneme substitution mistakes are more egregious than others. For example, a substitution of  {\ipa /p/} with {\ipa /b/} is better than a substitution of {\ipa /p/} and {\ipa /l/}, since the former pair only differs by one phonological feature (voicing). Finally, B-Cubed F-Score (BCFS) \citep{List2019} is a recently introduced metric that measures structural differences rather than substantial differences (i.e. no penalty if one phoneme is consistently substituted for another, but penalty for different predictions to the same ground truth), which is better suited for the supervised reconstruction task. 
Although these three other metrics exist, we found that they correlate rather highly with edit distance in most cases. As a result, we still focus our comparison of models on edit distance. 

\subsection{Baseline Methods}
There are two naive baseline methods that we can use to understand the boundary of the evaluation metrics. Without modeling proto-form reconstruction as a machine learning problem and using the reference proto-forms for supervised training, one baseline method is to select a \textit{random daughter} as the proto-form. This assumes that no sound changes occurred to differentiate the daughter languages from the proto-language. Another baseline to consider is the \textit{majority constituent} baseline, where the each of the existing daughter forms in a cognate set is segmented into constituents (onset, nucleus, and coda). The predicted proto-form is the concatenation of the most popular phoneme sequence of each constituent (onset, nucleus, and coda) among the daughter languages. This baseline relies on the syllabic structure of Sinitic languages, and it is based on the \textit{majority wins} heuristic from historical linguists \cite{campbell2013historical}. The results of these methods on Wikihan is shown in \autoref{tab:baselines}.

\begin{table*}[h!]
\centering
\begin{tabular}{@{}lrrrrr@{}}
\toprule
\textbf{Method}               & \multicolumn{1}{l}{\textbf{ED ↓}} & \multicolumn{1}{l}{\textbf{Acc \% ↑}} & \multicolumn{1}{l}{\textbf{PER ↓}} & \multicolumn{1}{l}{\textbf{FER ↓}} & \multicolumn{1}{l}{\textbf{BCFS ↑}} \\ \midrule
Random daughter      & 3.1181                              & 3.68                              & 0.7444                    & 0.2626                    & 0.3253                             \\
Majority constituent & 2.9187                              & 4.65                              & 0.7447                    & 0.2256                    & 0.4103                             \\ \midrule
CorPaR \citep{list-etal-2022-new}              & 1.2480                              & 32.04                             & 0.3435                    & 0.1509                    & 0.5697              \\
CorPaR + SVM \citep{list-etal-2022-new}        & 1.1940                              & 34.08                             & 0.3287                    & 0.1414                    & 0.5809              \\
Transformer \citep{11711-report}    & 0.9096                              & 52.73                             & 0.1878                    & 0.0740                    & 0.7206                             \\
RNN \citep{meloni-etal-2021-ab} & \textbf{0.8345}                              & \textbf{55.48}                             & \textbf{0.1719 }                   & \textbf{0.0680}                    & \textbf{0.7431}         \\ \bottomrule                  
\end{tabular}
\caption{Performance on various metrics of baseline methods (top 2) and previous methods (bottom 4) on WikiHan \citep{chang-etal-2022-wikihan}}
\label{tab:baselines}
\end{table*}

\section{Data Augmentation} \label{sec:data-augmentation}
\subsection{Motivation}
With limited resources, it is common for comparative datasets to contain incomplete cognate sets. This is also seen in the WikiHan dataset, where 93\% of the cognate sets have missing daughter forms, and the number of existing entries for each daughter language are imbalanced.
Thus it would be challenging to train neural models without overfitting on the small amount of data, and the models might overemphasize the daughter languages that have more existing entries when reconstructing the proto-forms.

To mitigate these problems, we augmented the training set of the WikiHan dataset by predicting the missing entries based on the proto-form and optionally the existing daughter forms in the same cognate set. With the additional entries in the training data especially for the languages with less data, we hope that the models could perform better in proto-form reconstruction, and have less variance.

\subsection{Reflex Prediction with CNN}
The CNN model used for the image inpainting task was the winner of the reflex prediction SIGTYP 2022 shared task \cite{kirov2022mockingbird}, thus we used the model for WikiHan data augmentation.

\begin{figure}
    \centering
    \includegraphics[width=\columnwidth]{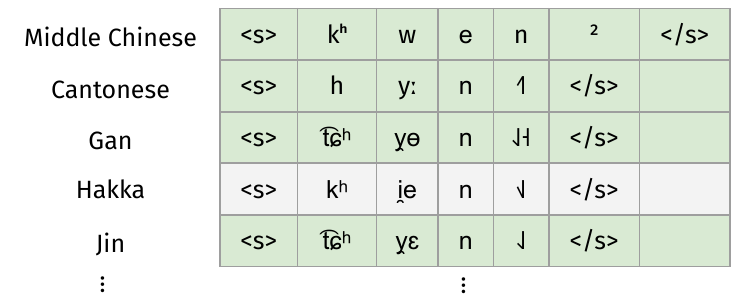}
    \caption{Stacked input for the CNN reflex prediction model.}
    \label{fig:augment-cnn}
\end{figure}

As shown in Figure \ref{fig:augment-cnn}, an entire cognate set is stacked as an input to the CNN model, where each row represents a language, and each column contains a phoneme. In this way, predicting a daughter form is similar to the image inpainting task, which recovers missing pixels of an image. The image inpainting CNN architecture mainly consists of a 2D convolution layer, dropout and non-linearity layers, and a deconvolution layer. While training, a random daughter form is masked and recovered as the deconvolution output to compute cross-entropy loss.

Since the method is sensitive to the number of existing/missing entries of each language, we create 10 random train/development splits, and train a CNN model for each of them. When applying the trained models to predict the missing entries in WikiHan, we ensemble the prediction results by taking the majority prediction among the 10 predicted daughter forms.

\subsection{Character-level Transduction}
Another data augmentation approach considers this problem in a different way. Instead of recovering a missing daughter entry from all the existing daughter entries and the proto-form, this approach only takes the proto-form and the language to predict as an input example.

\begin{figure}
    \centering
    \includegraphics[width=.9\columnwidth]{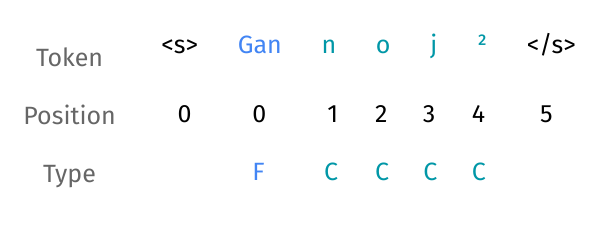}
    \caption{Example of position and type embeddings for feature-guided character-level transducer.}
    \label{fig:augment-transducer}
\end{figure}

This approach uses the transformer model proposed by \citeauthor{wu2021applying}, which has an character-level encoder-decoder architecture. The transducer takes additional features other than the characters as input. For the daughter form prediction use case, the feature input is the language to predict. In Figure \ref{fig:augment-transducer}, for the proto-form ``noj²'' and the daughter language Gan, the language is tokenized as a feature, and concatenated to the features of the proto-form characters. The position embedding marks the start of the characters, and the type embedding differentiates feature tokens from character tokens. The special embeddings are applied to avoid inconsistency caused by different relative distances from the feature to the characters.

\subsection{Experiments and Results}
We evaluate the data augmentation methods in two aspects. First, by masking existing daughter forms in the WikiHan dataset, we evaluate the accuracy of reflex prediction for both of the models. On top of that, we use the models to predict the actual missing entries of the WikiHan training data, and append the full cognate sets to the training data. With the additional cognate sets, we evaluate the proto-form reconstruction performance with the VAETransformer model introduced in the next section.

\subsubsection{Reflex Prediction}
The reflex prediction models are evaluated on the WikiHan dataset, and another Sinitic dataset, the Bai dataset from the SIGTYP 2022 shared task \cite{list2022sigtyp}, for reference. Compared with WikiHan where the languages are mostly from different linguistically well-established subgroups, the dialects in the Bai dataset are relatively closer both geographically (collected from different counties in the same province) and phonetically \cite{allen2007bai}. Since there is no proto-language in the Bai dataset, a random dialect is selected as the input form for the transducer model.

Table \ref{tab:reflex-pred} shows that the character-level transducer model significantly outperforms the CNN model on the WikiHan dataset for reflex prediction. On the contrary, CNN performed much better than the transducer on the Bai dataset.

A possible explanation is that the CNN image inpainting model learns from the correlations between neighboring ``pixels'', and for reflex prediction the neighbors include not only the neighboring phonemes for a reflex, but also the other languages in the cognate set. Therefore, the CNN model would learn to predict the missing reflexes more easily when the languages in a cognate set are more similar, which explains why it performed relatively well on Bai but poorly on WikiHan.

Meanwhile, the transducer is favorable when there are clear patterns to match from the input to the output phonemes. For the WikiHan dataset, the reflexes to predict are descended from the proto-form input, which make it natural to apply the transducer. On the other hand, without ancestral relationships between dialects of Bai, a CNN model that takes in entire cognate sets as input would be more preferable.

\begin{table}
\centering
\begin{tabular}{lcc}
\hline
\textbf{Method} & \textbf{WikiHan} (\%) & \textbf{Bai} (\%) \\
\hline
CNN & 38 & \textbf{55} \\
Transducer & \textbf{64} & 45  \\\hline
\end{tabular}
\caption{Reflex prediction dev set accuracy of both methods on the WikiHan dataset and the Bai dataset.}
\label{tab:reflex-pred}
\end{table}

\subsubsection{Proto-form Reconstruction}
With the WikiHan training data augmented with different methods, we train VAETransformer models with the same hyperparameters for proto-form reconstruction, and the performance is shown in Table \ref{tab:augment-result} \footnote{The numbers may differ from the numbers in the following section, because there are more hyperparameter tuning techniques applied for the experiments in the next section.}. Comparing the proto-form reconstruction performance with data augmentation via the CNN and transducer models, it is not surprising that a high-accuracy data augmentation model would result in better proto-form reconstruction. Also, data augmentation with transducer helped improving the edit distance, accuracy, and reduced the variance.

\begin{table}
\centering
\begin{tabular}{lcc}
\hline
\textbf{Method} & \textbf{Edit Distance} & \textbf{Accuracy} (\%) \\
\hline
None & $0.900 \pm 0.045$ & $51.6 \pm 2.31$\\
CNN & $0.935 \pm 0.017$ & $50.6 \pm 1.07$ \\
Transducer & $\textbf{0.870} \pm \textbf{0.015}$ & $\textbf{53.0} \pm \textbf{1.08}$   \\\hline
\end{tabular}
\caption{Proto-form reconstruction test set performance for the VAETransformer model with different data augmentation methods. Each method is trained with 3 runs to show the average and standard deviation of the metrics.}
\label{tab:augment-result}
\end{table}

\section{VAETransformer} \label{sec:vae-transformer}
\subsection{Motivation}
The daughter-in-proto-out architecture of Transformer or \citet{meloni-etal-2021-ab} is motivated by the nature of the reconstruction task. However, it ignores the one crucial aspect of the comparative method, namely the Neogrammarian hypothesis. This hypothesis claims that sound change is normally regular, meaning that a sound change rule has to apply to all words that meet the conditioning environment. This implies that there should be a mostly deterministic set of transformations from the proto form to arrive at the daughter forms, but the reverse is not true. For example, two proto-phonemes may merge into one phoneme in a daughter form, so it is impossible to recover the proto-phoneme by only looking at one daughter language\footnote{On the other hand, when one proto-phoneme splits into two phonemes, it is almost always conditioned by the neighboring environment, so the split is still a deterministic transformation.}. 
In the existing models, this hypothesis is not strictly enforced, but it is compensated by the fact that the input is a concatenation of all the available daughter forms, so the proto-form can still be predicted. We aim to improve the existing architecture by incorporating the Neogrammarian hypothesis in the form of adding a forward-reconstruction module. This is a model that predicts the daughter forms back from the proto form. It models the deterministic sound change rules that have occurred forward in time. The addition of this model will encourage the regularity of sound change in the model so that a better latent space is learned.

\begin{figure*}[h!]
    \centering
    \includegraphics[width=\textwidth]{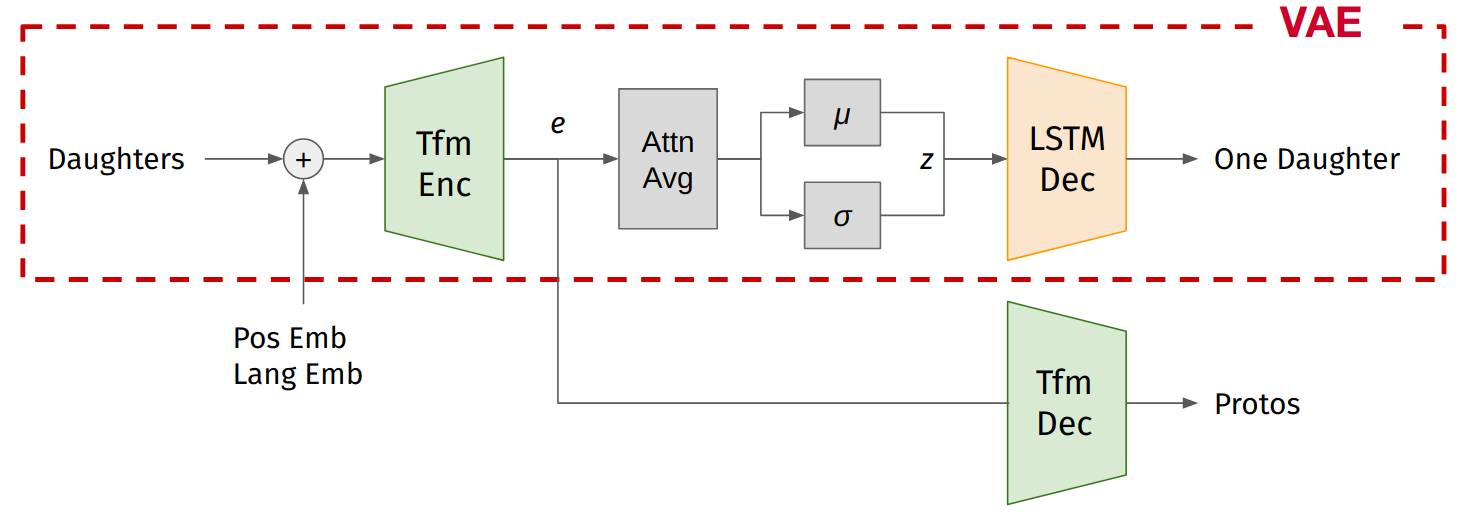}
    \caption{Model diagram of the proposed VAE-transformer model.}
    \label{fig:vae-transformer}
\end{figure*}

\subsection{Architecture}
The proposed model is shown in \autoref{fig:vae-transformer}. The model builds on top of the Transformer architecture (shown in green), with a transformer encoder and decoder to predict the proto form. However, in addition to the proto decoder, there is also a daughter decoder shown in orange to predict a daughter form back from the latent space. The daughter decoder is implemented as an LSTM model. 
Before the daughter decoder, there is an attention average block to transform the encoder output $e$ (which is a sequence of contexualized embeddings) into one latent vector. A Gaussian posterior distribution is then predicted from the latent space, and the reparameterization trick is used to produce the latent vector $z$, following the standard VAE architecture \citep{vae}. 

During training, all daughters get encoded to the latent space, but only one daughter language is randomly selected to be reconstructed\footnote{``Reconstruction" is used here in the autoencoder sense, not the historical linguistics sense.} from the latent space. We experimented with reconstructing all daughter forms during training instead of one, but the result was not as good.

The red dashed box represent the VAE part of the model. Compared to the Transformer model without VAE, the additional supervision encourages the model to learn a latent space ($z$ and ultimately $e$) that contains all the information to forward reconstruct any daughter language via a deterministic transformation. This should improve the quality of the latent space, and in turn, improve the performance on the proto-form reconstruction task.
\subsection{Experiments and Results}
Both VAE and Transformer are notorious for being sensitive to hyperparameters. When they are combined, the resulting model is not the easiest to tune. In this project, we explored both grid search and Bayesian search to find the optimal hyperparameters. In addition to the standard hyperparameters such as learning rate and batch size, there are four hyperparameters that contributed the most to the optimal performance:

\begin{itemize}[itemsep=0pt]
\item \textbf{Embedding Dimension} and \textbf{Feedforward Layer Dimension}: Since the dataset we work with is not considered large in the world in deep learning, the model size has to be carefully tuned to strike a balance between having enough expressive power and not overfitting to the dataset.
\item \textbf{Teacher Forcing Ratio}: In the daughter decoder, always using teacher forcing (i.e. feeding the ground truth token to the decoder regardless of the previously decoded token) during training can make the task too easy, especially for Sinitic where the word lengths are short. We find that disabling teacher forcing with some probability resulted in the best performance. 
\item \textbf{Warmup Epochs}: For transformer models, it is common to use a learning rate scheduler to gradually ramp up the learning rate. We use a linear warmup strategy, and we found that the performance can degrade if the warmup is too fast (5 epochs) or too slow (50 epochs). 
\end{itemize}

The full results are shown in \autoref{tab:results}. Other hyperparameters are shown in \autoref{appendix:hyperparameters}

\begin{table*}[h!]
\centering
\begin{tabular}{@{}lrrrrrr@{}}
\toprule
\textbf{Hyperparameter (default)}    & \multicolumn{1}{l}{\textbf{Value}} & \multicolumn{1}{l}{\textbf{ED ↓}} & \multicolumn{1}{l}{\textbf{Acc \% ↑}} & \multicolumn{1}{l}{\textbf{PER ↓}} & \multicolumn{1}{l}{\textbf{FER ↓}} & \multicolumn{1}{l}{\textbf{BCFS ↑}} \\ \midrule
emb size (128)              & 256                       & 0.9051                   & 52.44                        & 0.1878                    & 0.0705                    & 0.7166                     \\
                            & 64                        & 1.0240                   & 48.60                        & 0.2060                    & 0.0788                    & 0.6958                     \\ \midrule
feedforward size (64)       & 128                       & 0.9121                   & 51.94                        & 0.1894                    & 0.0728                    & 0.7158                     \\
                            & 96                        & 0.9004                   & 53.28                        & 0.1854                    & 0.0725                    & 0.7210                     \\
                            & 32                        & 0.8916                   & 53.50                        & 0.1854                    & 0.0725                    & 0.7212                     \\ \midrule
teacher forcing ratio (0.5) & 1.00                         & 0.9138                   & 52.50                        & 0.1884                    & 0.0737                    & 0.7192                     \\
                            & 0.75                      & 0.9258                   & 52.11                        & 0.1916                    & 0.0774                    & 0.7153                     \\
                            & 0.25                      & 0.8922                   & 53.24                        & 0.1844                    & 0.0735                    & 0.7245                     \\ \midrule
warmup epochs (15)          & 50                        & 0.9013                   & 52.69                        & 0.1850                    & \textbf{0.0700}                    & 0.7203                     \\
                            & 5                         & 0.8929                   & 52.05                        & 0.1861                    & 0.0720                    & 0.7214                     \\ \midrule
VAE-Transformer (default)                       & \multicolumn{1}{l}{}      & \textbf{0.8829}                   & \textbf{54.05}                        & \textbf{0.1822}                    & 0.0706                    & \textbf{0.7240}                     \\ \midrule
Transformer \citep{11711-report}  &  & 0.9096                              & 52.73                             & 0.1878                    & 0.0740                    & 0.7206                             \\
RNN \citep{meloni-etal-2021-ab} & & \textit{0.8345}                              & \textit{55.48}                             & \textit{0.1719 }                   & \textit{0.0680}                    & \textit{0.7431}         \\ \bottomrule
\end{tabular}
\caption{Performance of VAE-Transformer on various hyperparameter settings. A grid search experiment shows that model size, teacher forcing ratio, and warmup epochs are all important hyperparameters to tune in order to achieve the best performance. All results are the average of three or more runs. Bold figures are best in the hyperparameter search, and italic figures are the best overall.}
\label{tab:results}
\end{table*}

\section{Variational-NMT} \label{sec:variational-nmt}
\subsection{Motivation}
In the previous method, we adopted variational autoencoder (VAE) to reconstruct the daughter language from input daughter languages, in order to guide the generation of proto-languages. However, the goal of our model is still reconstructing the proto-form, the VAE loss enforced on daughter-to-daughter reconstruction is not a direct guide toward a better prediction.
Recall that VAE is designed to learn the underlying latent distribution $p_\theta(z)$ from a complicated input distribution $p_\theta(y)$. The lantent variable $z$ is learnt from encoder $q(z|y)$. During decoding, a standard VAE will learn a conditional likelihood distribution $p_\theta(y|z)$, and maximize the likelihood of data $x$ by $p_\theta(y) = p_\theta(y|z) p_\theta(y)$.
A VAE optimized on proto-language is capable of reconstructing plausible proto-forms. 
In this task, we do not want to reconstruct random proto-languages that do not have a corresponding historical basis. Instead, the reconstruction should be conditioned on the input daughter languages, which contain shared semantics for reconstruction. We utilized the idea of conditional VAE -- an extension of VAE that conditions the latent variable and data on additional variables. With an extra input $x$, the encoder is now conditioned on $y$ and $c$: $q(z|x, y)$ and the decoder becomes $p(y|z,x)$. In our specific task, the input distribution $y$ is proto-language and the condition $x$ is daughter languages.

\subsection{Architecture}
\begin{figure}[!htb]
    \includegraphics[width=0.5\textwidth]{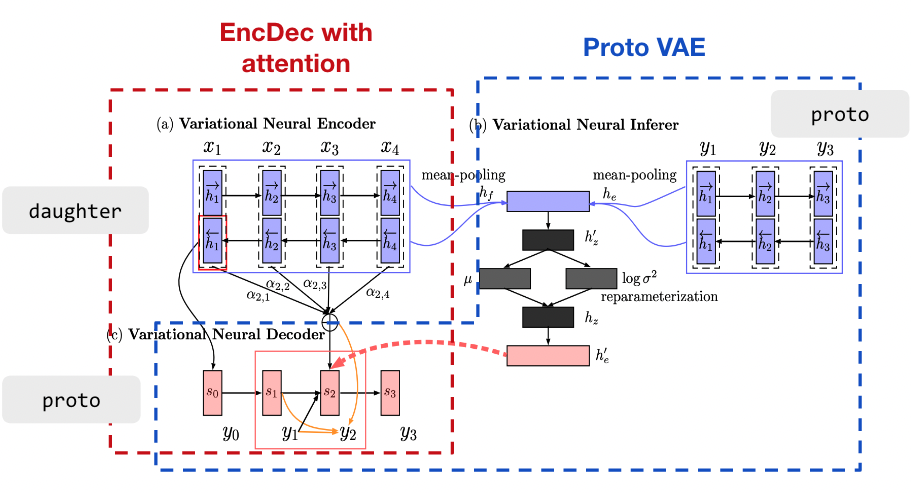}
    \caption{Model diagram of the proposed Variational-NMT model.}
    \label{fig:vnmt}
\end{figure}

\citet{Zhang2016VariationalNM} proposed a new structure variational neural machine translation (VNMT) to solve machine translation. They demonstrate that by introducing VAE for modeling the underlying semantics, VNMT can have a superior performance than an encoder-decoder structure with attention.
Inspired by this idea, we reimplement the VAE part on our GRU baseline \cite{meloni-etal-2021-ab}. Our model structure is shown in \autoref{fig:vnmt}. The model consists of two parts: 1) The encoder-decoder part, which is the same as Meloni's baseline; 2) The VAE part, which reconstructs proto-language by learning latent distributions. We will explain how we incorporated the VAE part into the encoder-decoder structure in detail.

In our model, except for a GRU encoder that encodes the concatenated cognate set into a hidden variable, we also have an encoder with the same structure but encodes the target proto-language instead. We extract the final hidden state $h_s$ and $h_t$ respectively from the two encoders. 
Using the same reparameterization trick proposed by VAE, we model the Gaussian parameters $\mu$ and $\sigma$ from two hidden states with neural networks and use them to obtain a representation of latent variable $z$:
\begin{align}
    h^{'}_{z} &= \text{act}(W_z[h_s;h_t]+b_z) \\
    \mu &= W_{\mu}h^{'}_{z}+b_{\mu} \\ 
    \text{log} \sigma^{2} &= W_{\sigma}h^{'}_{z}+b_{\sigma}
\end{align}
The sampling of $z$ from the distribution is converted to a deterministic dependency as $z = \mu + \sigma\odot\epsilon $, where $\epsilon \sim N(0, 1) $. Finally, latent variable $z$ is converted to the target hidden space by a linear layer, represented by $h_{e}$.

For the decoder input, apart from the encoder output weighted by attention, we also concatenate it with the latent variable $z$ as an extra input. The context vector $c_{j} $at timestamp $j$ is now represented as $ [ \sum_{i} \alpha_{ij}h_{i}; h_{e} ]$.

Our model is trained on reconstruction loss and KL divergence. The cross entropy loss is computed between the generated proto-language and the ground truth, which is the same loss for both VAE and GRU models. The KL divergence will enforce the two Gaussian distributions, prior $p(z|x)$ and posterior $p(z|x, y)$, to be close to each other. 

\subsection{Experiments}

We reimplement the above-mentioned VAE module on our GRU baseline \cite{meloni-etal-2021-ab}.
A hyper-parameter search is performed on the final implementation and the best result was shown in the table.
The dataset we used for this experiment is WikiHan.

\begin{table}
\centering
\begin{tabular}{lcc}
\hline
\textbf{Method} & \textbf{Edit Distance} & \textbf{Accuracy} (\%) \\
\hline
VNMT-prior & $0.874$ & $53.82$\\
VNMT & $0.855$ & $54.31$ \\
GRU & 0.843 & 54.50 \\\hline
\end{tabular}
\caption{Test set performance on Wikihan for the proposed VNMT model.}
\label{tab:vnmt-result}
\end{table}

\autoref{tab:vnmt-result} shows the experiment results. GRU refers to the baseline result from \cite{meloni-etal-2021-ab}. Our VNMT has a similar performance to the baseline.
We include the result of VNMT-prior, which is a variation of the model we mentioned above. When computing $ h^{'}_{z}$, instead of modeling with both hidden output from the two encoders, we used only the encoder of daughter languages, since the daughter and proto-languages are supposed to share the same semantic space. This method only considers the prior distribution $p(z|x)$ of the input cognate sets. The KL divergence is also different. We compute the divergence between the standard Gaussian distribution and the prior distribution.
In comparison, the standard VNMT version considers both prior and posterior distribution $p(z|x,y)$, since it models $ h^{'}_{z}$ from both daughter language and proto-language.
As is shown in the results, the VNMT with both prior and posterior terms has an edit distance of 0.85, which is lower than the version with only VNMT prior. This demonstrates the extra encoder for the proto-language helped to improve the reconstruction through the VAE.

\section{Analysis} \label{sec:analysis}
\subsection{Comparison with Previous Methods}
On the WikiHan dataset, VAE-Transformer out-performs the Transformer model \citep{11711-report} in every metric. We did notice that there is a higher variance among different runs when VAE is added. However, the difference is statistically significant when results are averaged over 8 runs: a one-sided \textit{t}-test on Edit Distance gives $p < 0.01 $. This means that adding the VAE structure improves the reconstruction quality. The daughter decoder enforces the latent space to follow the Neogrammarian hypothesis, i.e. it encourages the proto form latent space to contain all the information required to forward-reconstruct any single daughter form. This in turn produced a better reconstruction in the proto space. 

However, we note that VAE-Transformer does not out-perform the RNN model in \citet{meloni-etal-2021-ab}. One reason could be that this is a dataset-specific issue, since \citet{11711-report} demonstrated that the transformer model out performs RNN on the romance dataset and the Chinese dataset \citep{Hou2004}. RNN did not perform well on these because the sequence lengths are long, whereas WikiHan has shorter concatenated sequences than these two datasets. One next step is to do more rigorous analysis on the performance of these models on different datasets, as well as the effect of sequence length on performance. 

Another natural next step is to extend the architecture of \citet{meloni-etal-2021-ab} to add the VAE structure. If the VAE structure is universally effective in improve the latent space and reconstruction quality, it should further improve the performance of this state-of-the-art. However, when we conducted preliminary experiments with this idea, the VAE-RNN model instead performed worse than \citet{meloni-etal-2021-ab}. A more thorough analysis of the underlying causes is required and this is left as future work. 

\subsection{Prediction Errors}
\autoref{tab:error-analysis} shows the first five prediction mistakes the model makes in the test set, so the list is uncurated and gives a sense of the model's capability. In the first four rows, the model makes either substitution mistakes with similar phonemes or tones, or inserts an extra glide {\ipa /j/}. In the last row, the ground truth has an unusual nucleus due to the romanization system, so the model is unable to learn it since there is little reference for it in the dataset. Overall, the model seems to make sensible and understandable mistakes, and it does not behave too unexpectedly at least on WikiHan.

\begin{table}[h!]
\centering
\begin{tabular}{@{}ll@{}}
\toprule
Ground Truth & Prediction\\ \midrule
\ipa{x{\color[HTML]{4285F4}wo}n$^1$} & \ipa{x{\color[HTML]{CC002B}ju}n$^1$} \\
\ipa{ŋet$^4$} & \ipa{ŋ{\color[HTML]{CC002B}j}et$^4$} \\
\ipa{{\color[HTML]{4285F4}Ji}n$^2$} & \ipa{{\color[HTML]{CC002B}ne}n$^2$} \\
\ipa{meŋ$^{\color[HTML]{4285F4}2}$} & \ipa{meŋ$^{\color[HTML]{CC002B}1}$} \\
\ipa{k\super h{\color[HTML]{4285F4}j\textbari j}$^2$} & \ipa{k\super h{\color[HTML]{CC002B}i}$^2$} \\ \bottomrule                    
\end{tabular}
\caption{Uncurated examples of model prediction error.}
\label{tab:error-analysis}
\end{table}

\subsection{Attention Visualization}
\begin{figure}[h!]
    \centering
    \includegraphics[width=0.3\textwidth]{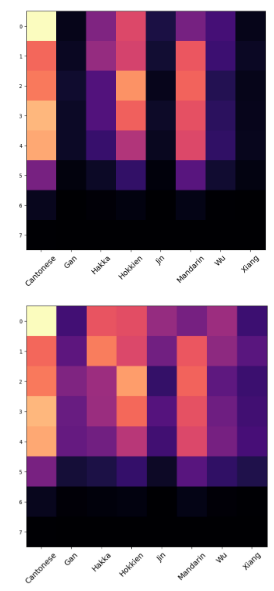}
    \caption{Attention weights on the different daughter languages. Weights are unnormalized (left) and normalized by the number of data points (right).}
    \label{fig:attention}
\end{figure}

We visualize the attention weights of the transformer encoder in order to understand what the model is extracting from the input. As shown in \autoref{fig:attention}, the attention on languages is imbalanced. Most of the weights fall on thre languages: Cantonese, Hokkien, and Mandarin. This makes sense because these are the three languages with the least amount of missing data in WikiHan. Even after the attention weights are normalized by the number of data points, these three languages are still most attended to. As a result, under this unbalanced dataset, the prediction can be biased towards the languages with most data. Trying to leverage all the daughter languages to produce the best reconstruction is left as future work. 

\section{Conclusions and Future Work} 
In this project, we set out to improve the performance of proto-language reconstruction on the WikiHan dataset. We found that data augmentation with a good reflex prediction model helps stabilize the model, yielding a mild improvement. Adding a VAE structure to the Transformer model also proved useful, although the model cannot beat \citet{meloni-etal-2021-ab} in the evaluation metrics. We also experiment with VAE ideas from the neural machine translation task and get comparable results to the baseline. When we examine the predictions from the model, we find that the mistakes are largely explainable linguistically, meaning that the model has learned phonetic knowledge implicitly. 

In terms of future work, more effort can be dedicated to adding a VAE structure to the RNN model \cite{meloni-etal-2021-ab}, since we have shown that VAE can improve the performance and the RNN model is currently state-of-the-art. Finally, we could also look into ways to improve the utilization of all daughter forms, so that more information can be leveraged to reconstruct the proto-form.

\bibliographystyle{acl_natbib}
\bibliography{acl2020}

\appendix

\section{Hyperparameters of VAETransformer} \label{appendix:hyperparameters}
\begin{table}[h!] 
\centering
\begin{tabular}{@{}ll@{}}
\toprule
Hyperparameter      & Value \\ \midrule
Learning rate      & 0.0015   \\
Batch size         & 64      \\
Adam $\beta_1$       & 0.9     \\
Adam $\beta_2$       & 0.98    \\
Dropout            & 0.07     \\
\# attention heads  & 8       \\
\# tfm encoder layers   & 4      \\
\# tfm decoder layers   & 4      \\ 
\# lstm decoder layers  & 1   \\
\# warmup epochs    & 15      \\ 
\# total epochs    & 200      \\ 
Embedding dimension    & 128      \\ 
Feedforward dimension    & 64      \\ 
KL loss multiplier      & 0.0015   \\
Teacher forcing ratio   &  0.5 \\ \bottomrule
\end{tabular}
\caption{Model configuration hyperparameters.}
\label{tab:model-config}
\end{table}

\end{document}